\documentclass[letterpaper,10pt,conference]{ieeeconf}
\IEEEoverridecommandlockouts
\usepackage[T1]{fontenc}
\usepackage{amsmath,amssymb}
\usepackage{booktabs}
\usepackage{graphicx}
\usepackage{array}
\usepackage{cite}
\usepackage{url}
\usepackage{flushend}
\usepackage[hidelinks]{hyperref}

\graphicspath{{figures/}}

\title{\LARGE \bf
FOCAL-VLA: Subtask-Guided Geometry Distillation and\\
Implicit World Modeling for Vision--Language--Action Models
}

\author{Zhiyuan Gao$^{1}$, Di Wen$^{2}$, Yanxiang Zhan$^{1}$, Mohammad Khoshnazar$^{1}$,\\
Jeroen Sch\"afer$^{1}$, Kunyu Peng$^{2}$, Michael Beetz$^{1,3}$%
\thanks{\raggedright $^{1}$University of Bremen, Germany.\newline
Emails: \mbox{gao@uni-bremen.de},
\mbox{yanxiang@uni-bremen.de},
\mbox{khoshnam@uni-bremen.de},
\mbox{jeroen.schaefer@uni-bremen.de},
\mbox{beetz@informatik.uni-bremen.de}.}%
\thanks{\raggedright $^{2}$Karlsruhe Institute of Technology (KIT), Germany.\newline
Emails: \mbox{di.wen@kit.edu}, \mbox{kunyu.peng@kit.edu}.}%
\thanks{\raggedright $^{3}$Robotics Institute Germany (RIG).}}

\begin{document}
\bstctlcite{BSTcontrol}

\maketitle
\thispagestyle{empty}
\pagestyle{empty}
\raggedbottom

\begin{abstract}
Vision--language--action (VLA) models built on pretrained vision--language
models have demonstrated strong performance across diverse robotic
manipulation tasks.  However, VLA models that directly map current 2D
observations to actions often lack sufficient spatial and temporal
understanding, limiting their performance in precise and long-horizon
manipulation.  Recent methods enhance VLA models through geometric supervision
and future-state prediction across the entire scene.  However, these methods
can suffer from redundant scene information, distracting the model from
learning the geometry and dynamics relevant to the current interaction.
To address this issue, we propose
FOCAL-VLA, a framework that combines subtask-guided geometry distillation
with implicit world modeling to learn representations of current spatial
structure and future interaction dynamics.
To focus geometric learning on the current subtask, we transfer
geometric knowledge from VGGT to the VLA model by aligning geometry latents
with features from subtask-relevant image regions.
To capture the future 3D evolution of the current interaction, we incorporate
implicit world modeling using Track4World features from
current and future demonstration frames.
The two complementary representations
jointly guide action generation without running VGGT or Track4World at inference
time.  Experiments show that FOCAL-VLA outperforms baselines on both
simulation benchmarks and real-world
manipulation tasks.
Project website: {\urlstyle{same}\url{https://zhiyuan-gao.github.io/FOCAL-VLA/}}.
\end{abstract}

\begin{figure*}[t]
    \centering
    \includegraphics[width=\textwidth]{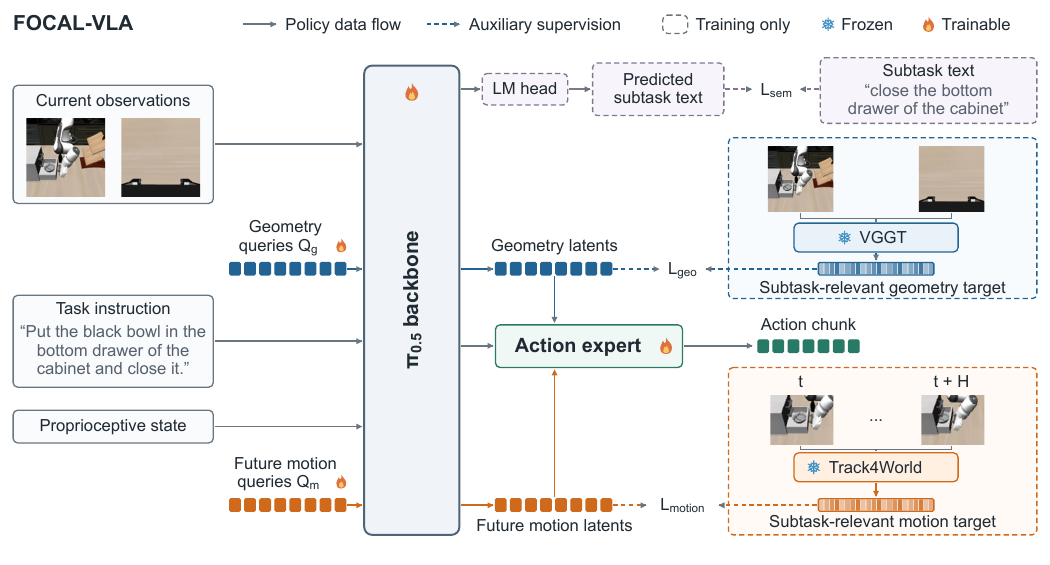}
    \caption{Overview of FOCAL-VLA.  The $\pi_{0.5}$ backbone processes current
    observations, the task instruction, proprioceptive state, and two groups of
    learnable queries.  The output hidden states at the query positions form
    geometry and future motion latents, which jointly condition the action expert
    alongside the native vision--language context.  Geometry distillation and
    implicit world modeling use cached, subtask-relevant targets constructed
    from frozen VGGT and Track4World features, respectively
    (Fig.~\ref{fig:stage_targets}).  An auxiliary language-modeling branch
    predicts subtask text through the native LM head.  Dashed borders indicate
    training-only branches; distillation prediction heads are omitted for clarity.
    The motion thumbnails show enlarged regions at $t$ and $t+H$, where $H$ is
    the action horizon; Track4World receives complete frames.}
    \label{fig:overview}
\end{figure*}

\section{Introduction}
\label{sec:introduction}

Vision--language--action (VLA) models extend pretrained vision--language
models to robotic control, enabling robots to perform diverse tasks and
generalize to new objects and environments
\cite{brohan2023rt2,kim2024openvla,black2024pi0,black2025pi05}.
Advances in large-scale robot learning
\cite{brohan2022rt1,ghosh2024octo} and training across embodiments
\cite{oneill2024openx} have further expanded these capabilities.
However, precise and long-horizon manipulation requires more than semantic
understanding: robots must perceive the spatial relationships between
interacting entities and anticipate how an interaction will evolve
\cite{li2026spatialforcing,zhang2025dreamvla,lin2025hifvla}.
To strengthen spatial understanding, recent methods incorporate 3D
representations and spatial encodings into VLA models
\cite{zhen2024threedvla,qu2025spatialvla,zhang2026falcon}, or transfer
geometric knowledge from pretrained 3D foundation models through
representation alignment
\cite{li2026spatialforcing,guo2025glad,shi2026threedthinkvla}.
The latter provides geometric supervision during training without requiring
the teacher at inference time.  A complementary direction improves temporal
understanding by incorporating motion history or predicting future
observations \cite{zheng2025tracevla,du2023unipi,wu2023gr1}.
These approaches range from visual trace prompting
\cite{zheng2025tracevla} and joint image--action prediction
\cite{wu2023gr1} to learning future visual features, 3D point tracks, and
scene flow \cite{zhang2025dreamvla,kim2026pri4r,wang2026lamp,wang2026track4action}.
Together, these developments motivate combining representations of current
spatial structure with those of future interaction dynamics.

The usefulness of spatial and temporal information, however, depends on its
relevance to the current interaction.  Recent studies address this
requirement through visual attention conditioned on instructions and
execution history \cite{xiao2026avavla}, or through structured affordance
representations and progress estimation \cite{liu2026palm}.
For geometry and future-motion distillation, the spatial scope of the
supervisory targets is therefore an important design choice.
Existing methods that align all visual tokens
\cite{li2026spatialforcing,guo2025glad} or pool motion features across the
entire scene \cite{wang2026track4action} also supervise regions unrelated
to the current interaction.  Such supervision can introduce redundant scene
information, distracting the model from learning the geometry and dynamics
relevant to the current operation.  Moreover, the information needed for
action generation changes as the interaction progresses.  For example,
placing an object into a drawer requires understanding the object--drawer
spatial relationship, whereas closing the drawer requires anticipating the
drawer's motion.  These observations motivate constructing geometric and
future-motion supervision around the entities involved in the current
interaction.

To address these limitations, we propose FOCAL-VLA, built on
$\pi_{0.5}$~\cite{black2025pi05}, which combines subtask-guided geometry
distillation with implicit world modeling to guide action generation
(Fig.~\ref{fig:overview}).  To focus geometric learning on the
current subtask, we distill features from frozen VGGT~\cite{wang2025vggt}
into geometry latents by aggregating teacher features only from regions
corresponding to the entities involved in that subtask.  The selected regions
change with the interacting entities across subtasks.  To anticipate how the
current interaction will unfold, we incorporate implicit world modeling through
future motion latents learned from current observations.
Frozen Track4World~\cite{lu2026track4world} extracts features from current and
future demonstration frames.  We aggregate features only from regions relevant
to the current subtask to supervise the future motion latents.  The two
complementary representations jointly condition the action expert, without running
either teacher at inference time.
The subtask annotations and relevance masks used to construct distillation
targets are not required at inference time.

We evaluate FOCAL-VLA on LIBERO~\cite{liu2023libero},
RoboCasa~\cite{nasiriany2024robocasa}, and four real-world manipulation tasks.
FOCAL-VLA performs competitively with existing VLA baselines
on both simulation benchmarks and outperforms the $\pi_{0.5}$ baseline on
real-world manipulation tasks.
Ablation studies further demonstrate the complementary benefits
of geometry and future motion supervision, while showing that restricting
supervision to subtask-relevant regions outperforms full-image supervision.

\begingroup
\setlength{\parskip}{0pt}
In summary, our main contributions are:
\begin{itemize}[\setlength{\topsep}{0pt}%
    \setlength{\partopsep}{0pt}%
    \setlength{\parsep}{0pt}%
    \setlength{\itemsep}{0.15em}]
    \item To focus geometric learning on the current subtask, we transfer
    VGGT's geometric knowledge to the VLA model through distillation over
    subtask-relevant regions, adapting the selected regions as the interacting
    entities change.
    \item To anticipate how the current interaction will unfold, we introduce
    implicit world modeling through future motion latents, using future
    demonstration context while restricting spatial supervision to regions
    relevant to the current subtask.
    \item To evaluate the effectiveness of FOCAL-VLA, we conduct experiments
    on simulation benchmarks and real-world manipulation tasks, complemented
    by ablation studies examining the contributions of the distillation
    branches, supervision scope, and action conditioning.
\end{itemize}
\endgroup

\section{Related Work}
\label{sec:related_work}

\subsection{Vision--Language--Action Models}
VLA models adapt pretrained vision--language representations to robot action
generation~\cite{brohan2023rt2,kim2024openvla,black2024pi0}.  Representative
designs include autoregressive action tokenization in
OpenVLA~\cite{kim2024openvla} and continuous flow-matching action generation
in $\pi_0$~\cite{black2024pi0}.  Recent generalist models further expand
capabilities for instruction following and transfer across tasks and embodiments:
$\pi_{0.7}$~\cite{physicalintelligence2026pi07} uses diverse context
conditioning to learn from heterogeneous data, while
LingBot-VLA~\cite{wu2026lingbotvla} scales pretraining across multiple
dual-arm robot configurations.  Our base policy,
$\pi_{0.5}$~\cite{black2025pi05}, combines a continuous action expert with
heterogeneous co-training that includes semantic subtask prediction.
FOCAL-VLA builds on this architecture and augments its policy representations
through subtask-guided geometry distillation and implicit world modeling.

\subsection{Geometry-Enhanced VLA Models}
One approach to improving spatial understanding is to provide the policy with
explicit geometric information, including point-cloud embeddings
\cite{li2025pointvla,sun2025geovla}, robot-centric pointmaps
\cite{lee2026pointmap}, and egocentric 3D position encodings
\cite{qu2025spatialvla}.  Rather than supplying explicit geometric inputs to
the policy, feature-distillation methods transfer geometric knowledge from
pretrained 3D foundation models into policy representations without requiring the teacher
at inference time.  Spatial Forcing~\cite{li2026spatialforcing} and
GLaD~\cite{guo2025glad} align a single VLA backbone layer with VGGT features,
while ROCKET~\cite{sun2026rocket} extends alignment to multiple layers
through a shared projector.
3DThinkVLA~\cite{shi2026threedthinkvla} incorporates both geometric and
spatial reasoning features into action-query states to guide action prediction.
These methods align geometric features across the entire image rather than
focusing the alignment on regions relevant to the current subtask.
We align the policy's geometry latents with teacher features from
subtask-relevant regions, adapting these regions as the interacting entities
change across subtasks.
The resulting geometry latents directly condition the action expert.

\subsection{Future Prediction for VLA Models}
One line of work guides VLA action generation by predicting future images
or visual features.
CoT-VLA~\cite{zhao2025cotvla} generates a future goal image before generating
its action chunk, while DreamVLA~\cite{zhang2025dreamvla} predicts future dynamic
regions, depth, and semantic features.  Related world-action models couple
future video generation with action prediction or planning
\cite{kim2026cosmospolicy,lopetegui2026sawam}.

Other methods represent future changes through motion features or 3D
trajectories.  CoWVLA~\cite{yang2026cowvla} learns a factorized motion
representation from video, while GeoPredict~\cite{qian2026geopredict}
combines future robot-track prediction with predictive 3D Gaussian geometry.
Pri4R~\cite{kim2026pri4r} uses auxiliary future 3D point-track prediction
during training while retaining the original VLA architecture at deployment.
LaMP~\cite{wang2026lamp} learns future 3D scene flow with a motion expert
and supplies its denoising features to the action expert.
Track4Action~\cite{wang2026track4action} pools scene and motion features from
Track4World over an action-aligned clip into a tracker target that supervises
query states.  These query states condition action generation through gated fusion.
Our method uses the same teacher but restricts feature distillation to regions
relevant to the current subtask.  This supervises the policy's future motion
latents using features that capture upcoming 3D changes in the interacting entities.

\section{Method}
\label{sec:method}

In this section, we present FOCAL-VLA.  We begin with the preliminaries
(Sec.~\ref{sec:method_overview}), followed by subtask-guided geometry
distillation (Sec.~\ref{sec:geometry}) and implicit world modeling through
future-informed motion distillation (Sec.~\ref{sec:motion}).  We then describe action generation and training
(Sec.~\ref{sec:action_conditioning}).  An overview of FOCAL-VLA is shown in
Fig.~\ref{fig:overview}.

\subsection{Preliminaries}
\label{sec:method_overview}

We first review the VLA formulation using $\pi_{0.5}$~\cite{black2025pi05},
which serves as our base policy.  It combines a pretrained PaliGemma
vision--language backbone with a continuous action expert.  At time $t$,
the backbone encodes the current multi-view observation $o_t$, task
instruction $\ell$, and tokenized proprioceptive state into context tokens
$C_t$.  Conditioned on $C_t$, the
action expert predicts an action chunk
$a_{t:t+H-1}=(a_t,\ldots,a_{t+H-1})$, where $a_t$ denotes a robot action
and $H$ is the prediction horizon.  The action expert is trained on expert
demonstrations through conditional flow matching and generates actions by
integrating a learned velocity field from Gaussian noise.

Building on this architecture, FOCAL-VLA introduces two groups of learnable
queries, $Q_g$ and $Q_m$, as additional input tokens to the vision--language
backbone.  Their output hidden states form the geometry latents $Z_g$ and
future motion latents $Z_m$, respectively, with the attention pattern defined
in Sec.~\ref{sec:action_conditioning}.
The geometry latents are aligned with subtask-relevant VGGT features, while
the future motion latents are aligned with future-informed Track4World
features.  Both groups provide additional conditioning to the action expert
alongside $C_t$.

\subsection{Subtask-Guided Geometry Distillation}
\label{sec:geometry}

\begin{figure*}[t]
    \centering
    \includegraphics[width=\textwidth]{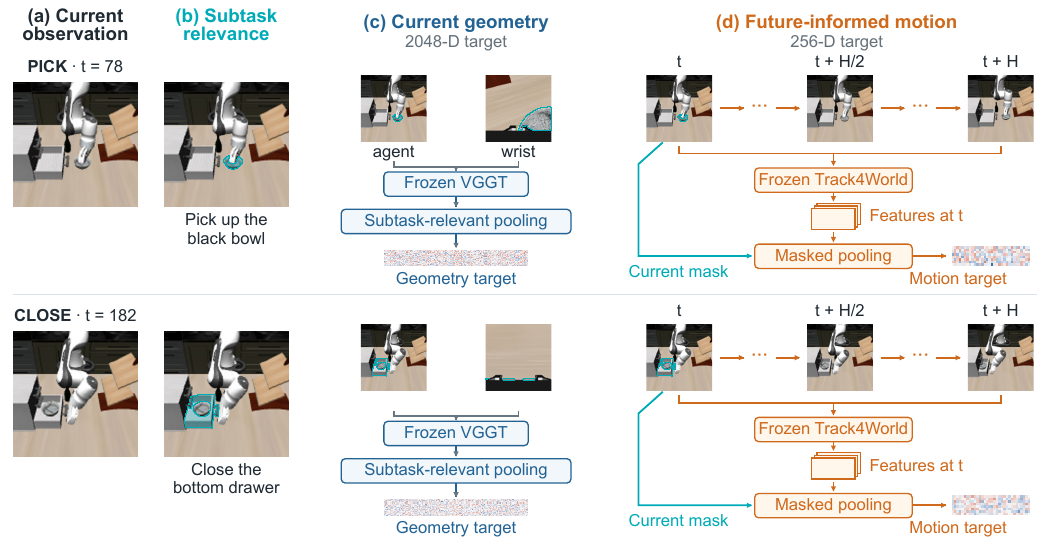}
    \caption{Construction of subtask-relevant geometry and future motion
    targets.  (a--b) PICK and CLOSE from the same demonstration select different
    entities: the black bowl and the bottom drawer, respectively.  (c) VGGT
    features are pooled within the selected regions in each current view and
    averaged across valid views to form the geometry target.  (d) Track4World
    processes current and future frames over the action horizon $H$.  The
    current subtask mask selects features at time $t$, enriched by future
    context, for motion-target pooling.  Mask overlays indicate pooling regions;
    both teachers receive complete images.  Color blocks show standardized
    cached target vectors on a shared $[-4,4]$ scale, not spatial maps;
    feature stacks are schematic.}
    \label{fig:stage_targets}
\end{figure*}

Existing geometry-distillation methods
align VLA visual embeddings with 3D foundation-model features across the
entire image.  This
alignment includes scene regions unrelated to the current operation and does
not account for changes in the relevant entities across subtasks.  To focus
geometric learning on the current interaction, we propose subtask-guided
geometry distillation, which aligns the policy's geometry latents with
features from our geometry teacher, VGGT~\cite{wang2025vggt}, aggregated over
entities involved in the current subtask
(Fig.~\ref{fig:stage_targets}).  The geometry query group $Q_g$ consists of
eight learnable tokens appended to $C_t$.

VGGT is a feed-forward multi-view model for camera and scene geometry
estimation.  The frozen model jointly
processes the current-view images, yielding final-layer patch features
$G_t^v\in\mathbb{R}^{P_g\times d_g}$ for each view $v\in\mathcal{V}$,
where $P_g$ is the number of patches per view and $d_g=2048$.  To select features relevant to the current subtask, we
use an offline vision--language model to align manually specified subtask
descriptions with demonstration frames and identify the entities involved.
A binary relevance mask $M_t^v$
is formed by taking the union of the whole-entity masks for entities involved
in the current subtask.  Let
$w_{t,p}^v\in[0,1]$ denote the coverage of the transformed relevance mask
on patch $p$.  We aggregate features only within these subtask-relevant
regions, pooling within each view and averaging over views with nonzero
mask coverage:
\begin{equation}
    y_t^g = \frac{1}{|\mathcal{V}_t|}
    \sum_{v\in\mathcal{V}_t}
    \frac{\sum_{p=1}^{P_g} w_{t,p}^v G_{t,p}^v}
         {\sum_{p=1}^{P_g} w_{t,p}^v},
    \label{eq:geometry_target}
\end{equation}
where $\mathcal{V}_t=\{v:\sum_p w_{t,p}^v>0\}$ is the set of valid views.

A prediction head applies per-query normalization, mean pooling, and a linear
projection to produce $\hat{y}_t^g=h_g(Z_g)\in\mathbb{R}^{d_g}$.
For each branch $b\in\{g,m\}$, we standardize targets and predictions using
$\widetilde{x}=(x-\mu_b)/\sigma_b$, where $\mu_b$ and $\sigma_b$ are the
per-dimension target mean and standard deviation computed on the training split.
We align the geometry latents with the pooled VGGT target using a masked
mean-squared error.  Let $\mathcal{B}_g$ contain the batch samples with at least
one valid view:
\begin{equation}
    \mathcal{L}_{\mathrm{geo}} =
    \frac{1}{|\mathcal{B}_g|d_g}
    \sum_{t\in\mathcal{B}_g}
    \left\|\widetilde{\hat{y}}_t^g-
    \widetilde{y}_t^g\right\|_2^2.
    \label{eq:geometry_loss}
\end{equation}

\subsection{Implicit World Modeling}
\label{sec:motion}

We implement implicit world modeling through future-informed motion
distillation.  Conditioned on the current context $C_t$, the model learns
to infer a latent representation of upcoming 3D interaction dynamics,
using future demonstration frames only to construct training targets.
Recent approaches
incorporate future motion representations into VLA models.  However,
full-scene feature pooling~\cite{wang2026track4action} combines motion cues
from interacting entities with features from regions unrelated to the current
subtask.
Our future-informed motion distillation addresses this issue by restricting
feature aggregation to the currently relevant entities while retaining
temporal context from future demonstration frames.  Aggregated features from
our motion teacher, Track4World~\cite{lu2026track4world}, provide targets for
the policy's future motion latents
(Fig.~\ref{fig:stage_targets}).  We implement this
branch with eight learnable query tokens $Q_m$ appended to $C_t$.

Track4World is a feed-forward model that estimates dense 3D scene flow
between video frames and reconstructs point trajectories in a shared world
coordinate system.  We use its frozen motion features, rather than the
reconstructed trajectories, as distillation targets.  By processing current
and future demonstration frames together, the teacher provides motion
features informed by the subsequent interaction.
For a training anchor at time
$t$, let
$\mathcal{T}_t=\{\tau_k\}_{k=1}^{K}$ denote future timestamps
sampled uniformly over the demonstration interval corresponding to the next
action chunk, excluding the current frame.  Sampling spans the full chunk
interval without truncation at subtask boundaries.  The teacher processes the
current agent-view frame
$o_t^{\mathrm{agent}}$ together with
$\{o_{\tau_k}^{\mathrm{agent}}\}_{k=1}^{K}$, for $K+1$ frames in total.
The current agent-view mask $M_t^{\mathrm{agent}}$ selects the features used
for distillation.

Let $D_{t,p}\in\mathbb{R}^{d_m}$ denote the final output of Track4World's
temporally global 3D flow-feature aggregator at source time index zero and
source patch $p$, after the teacher has processed the full clip, where
$d_m=256$.  Mapping the current relevance mask to the teacher's
$45\times45$ source grid ($P_m=2025$) gives coverage weights $u_{t,p}$.  The motion target
is
\begin{equation}
    y_t^m =
    \frac{\sum_{p=1}^{P_m}u_{t,p}D_{t,p}}
         {\sum_{p=1}^{P_m}u_{t,p}}.
    \label{eq:motion_target}
\end{equation}
Each $D_{t,p}$ incorporates temporal context from the full clip, while the
current mask determines which entities contribute to $y_t^m$.  Aligning
$Z_m$ with this target trains the policy to infer future motion features of
the current subtask's entities from its current context.
Future evolution is thus modeled in latent feature space, without explicitly
reconstructing future images or 3D trajectories.

A separate head with the same normalize--mean--project structure predicts
$\hat{y}_t^m=h_m(Z_m)\in\mathbb{R}^{256}$, supervised by a masked Smooth L1
loss with transition parameter $\beta=1$ in the standardized feature space:
\begin{equation}
    \mathcal{L}_{\mathrm{motion}} =
    \frac{1}{|\mathcal{B}_m|}
    \sum_{t\in\mathcal{B}_m}
    \operatorname{SmoothL1}_{\beta=1}
    \left(
      \widetilde{\hat{y}}_t^m,
      \widetilde{y}_t^m
    \right).
    \label{eq:motion_loss}
\end{equation}
Here Smooth L1 is averaged over the feature dimensions, and
$\mathcal{B}_m$ contains batch samples with a valid current agent mask and
all $K$ sampled future frames.

\subsection{Action Generation and Training}
\label{sec:action_conditioning}
\label{sec:objective}
\label{sec:semantic}

The geometry and future motion query groups attend independently to $C_t$
and to tokens within their own groups.  The action expert attends to both
groups alongside $C_t$, while the context tokens retain their native pathway.
The attention pattern is
\begin{equation}
\begin{aligned}
    C_t &\leftarrow C_t,\\
    Z_g &\leftarrow C_t + Z_g,\\
    Z_m &\leftarrow C_t + Z_m,\\
    \text{Action} &\leftarrow C_t + Z_g + Z_m.
\end{aligned}
\label{eq:attention_topology}
\end{equation}
Here each left-hand side denotes the token group being updated, the right-hand
side lists the groups it may attend to, and $+$ denotes their union.
For the action expert, only external conditioning groups are shown;
its native internal attention pattern is retained.

To provide the shared context with subtask-level semantic guidance, we also
train the model's native language-modeling branch to predict the subtask
text obtained in Sec.~\ref{sec:geometry}.  This branch reuses the PaliGemma
tokenizer, token embeddings, transformer, and tied language-model head to
generate the text autoregressively from $C_t$.  Given target
tokens $s_{1:L_s}$ of length $L_s$, the semantic objective is
\begin{equation}
    \mathcal{L}_{\mathrm{sem}} =
    -\frac{1}{L_s}\sum_{k=1}^{L_s}
    \log p_\theta(s_k\mid C_t,s_{<k}),
    \label{eq:semantic_loss}
\end{equation}
including the end-of-sequence token and excluding padded positions.
The native autoregressive branch requires no additional semantic query or
semantic embedding head.  Its attention mask allows target tokens to attend
to $C_t$ and their causal prefix, but prevents the context, geometry, future
motion, and action tokens from attending to semantic target tokens.  The
semantic loss therefore trains the shared vision--language context without
providing ground-truth subtask text to the action pathway.

The overall objective is
\begin{equation}
    \mathcal{L} = \mathcal{L}_{\mathrm{action}}
    +\lambda_g\mathcal{L}_{\mathrm{geo}}
    +\lambda_s\mathcal{L}_{\mathrm{sem}}
    +\lambda_m\mathcal{L}_{\mathrm{motion}},
    \label{eq:total_loss}
\end{equation}
where $\mathcal{L}_{\mathrm{action}}$ is the native $\pi_{0.5}$ action
objective, and $\lambda_g$, $\lambda_s$, and $\lambda_m$ weight the three
auxiliary objectives.  We jointly optimize the policy parameters, query
embeddings, and prediction heads using the cached distillation targets.

At inference time, the policy computes $Z_g$ and
$Z_m$ from the current observation, instruction, and proprioceptive state
and uses them to condition
the action expert.  The teachers, future demonstration frames, relevance
masks, and semantic targets are not used.

\section{Experiments}
\label{sec:experiments}

We evaluate FOCAL-VLA on the LIBERO and RoboCasa simulation benchmarks
and on real-world robotic manipulation tasks.  We further conduct ablation
studies to examine the contributions of individual components and the effects
of our design choices.

\begin{table}[t]
    \centering
    \caption{Comparison on LIBERO (success rate, \%).
    Bold indicates the best result in each column.}
    \label{tab:libero}
    \small
    \setlength{\tabcolsep}{3pt}
    \begin{tabular*}{\columnwidth}{@{\extracolsep{\fill}}lccccc@{}}
        \toprule
        Method & Spatial & Object & Goal & Long & Avg. \\
        \midrule
        OpenVLA~\cite{kim2024openvla}
            & 84.7 & 88.4 & 79.2 & 53.7 & 76.5 \\
        $\pi_0$~\cite{black2024pi0}
            & 96.8 & 98.8 & 95.8 & 85.2 & 94.2 \\
        $\pi_{0.5}$~\cite{black2025pi05}
            & 98.8 & 98.2 & \textbf{98.0} & 92.4 & 96.9 \\
        OpenVLA-OFT~\cite{kim2025openvlaoft}
            & 97.6 & 98.4 & 97.9 & 94.5 & 97.1 \\
        GR00T N1.7~\cite{lerobot2026groot17}
            & 95.0 & \textbf{100.0} & \textbf{98.0} & 93.0 & 96.5 \\
        SpatialVLA~\cite{qu2025spatialvla}
            & 88.2 & 89.9 & 78.6 & 55.5 & 78.1 \\
        DreamVLA~\cite{zhang2025dreamvla}
            & 97.5 & 94.0 & 89.5 & 89.5 & 92.6 \\
        GeoPredict~\cite{qian2026geopredict}
            & 98.0 & 98.2 & 95.7 & 94.0 & 96.5 \\
        Track4Action~\cite{wang2026track4action}
            & 95.0 & 99.6 & 97.6 & \textbf{95.8} & 97.0 \\
        \midrule
        \textbf{FOCAL-VLA}
            & \textbf{100.0} & 98.6 & 97.8
            & 95.0 & \textbf{97.9} \\
        \bottomrule
    \end{tabular*}
\end{table}

\subsection{Experimental Setup}
\label{sec:exp_setup}

\paragraph{Simulation benchmarks}
We train one policy per benchmark.  LIBERO~\cite{liu2023libero} comprises
Spatial, Object, Goal, and Long (LIBERO-10), each with ten tasks; the policy
uses agent and wrist views and predicts and executes 10-step action chunks.
For RoboCasa~\cite{nasiriany2024robocasa}, we use Atomic-24 under Human-50
(50 demonstrations per task), with left agent, wrist, and right agent views;
the policy predicts 50-step chunks and executes 25 steps before replanning.
We evaluate 50 episodes per task, reporting LIBERO's per-suite success rates
and their unweighted average, and RoboCasa's unweighted average task success rate.

\begin{figure}[t]
    \centering
    \includegraphics[width=\columnwidth]{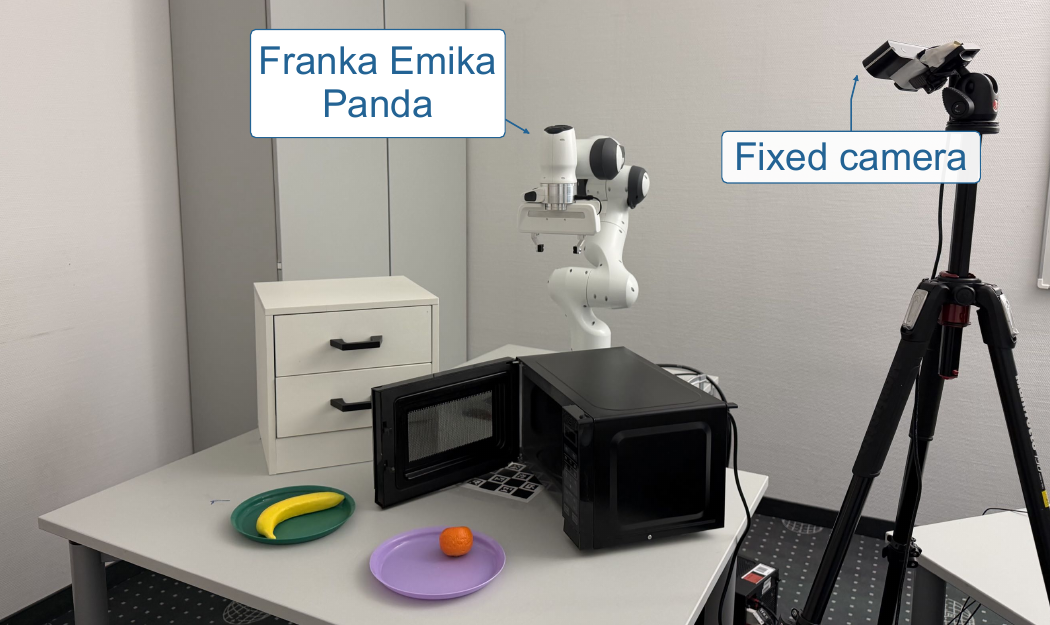}
    \caption{Real-world platform with a Franka Emika Panda and a fixed
    external camera.  The workspace contains the objects used across the
    four manipulation tasks.}
    \label{fig:real_world_setup}
\end{figure}

\begin{figure*}[t]
    \centering
    \includegraphics[width=\textwidth]{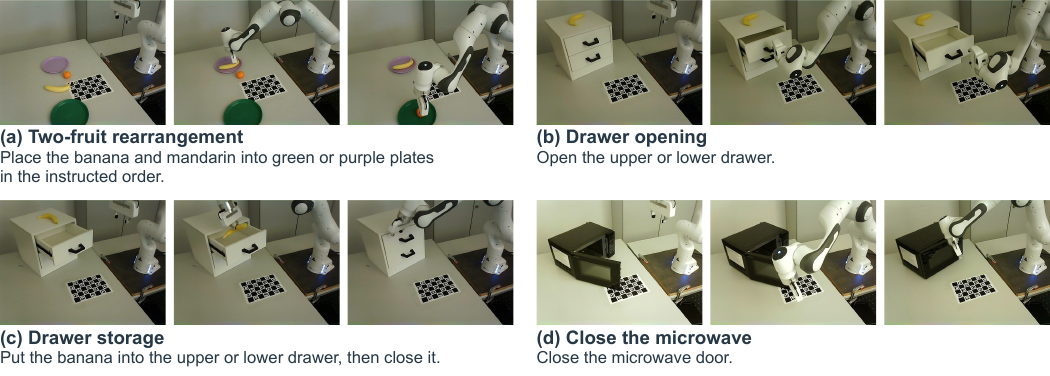}
    \caption{Illustration of the four real-world tasks.  Representative
    frames from each task are shown in temporal order from left to right,
    viewed from the fixed external policy camera.  Task descriptions appear
    below each sequence.  A uniform color correction is applied for display.}
    \label{fig:real_world_tasks}
\end{figure*}

\paragraph{Real-world experiments}
We use a Franka Emika Panda with one fixed external Orbbec Astra Pro RGB
camera (Fig.~\ref{fig:real_world_setup}).  Fig.~\ref{fig:real_world_tasks}
illustrates four tasks: sequentially placing a banana and mandarin into
instructed plates (possibly the same plate), fully opening a specified
upper or lower drawer, storing a banana in a specified initially open
upper or lower drawer and closing it, and closing the microwave.
We collect 50--150 successful demonstrations per task using SpaceMouse
teleoperation, totaling 350 demonstrations recorded at 10\,Hz with
$640\times480$ RGB observations.
FOCAL-VLA and $\pi_{0.5}$ each learn one policy across all four tasks,
predicting and executing 10-step chunks.  We report per-task and overall
success rates over 20 trials per task, using the same initial-layout
distribution for both methods.  Success requires completing all instructed
actions in order and reaching the specified final configuration.

\paragraph{Implementation details}
We fine-tune the official OpenPI $\pi_{0.5}$ implementation end to end.
Offline annotation (Sec.~\ref{sec:geometry}) uses
Qwen3-VL-8B-Instruct~\cite{qwen2025qwen3vl} to identify subtask starts and
entities from sampled demonstration frames, task instructions, and manually
specified subtask descriptions.  Entities include manipulated objects and
associated containers, supports, or articulated objects, excluding the gripper.
Masks come from simulator annotations or entity-prompted
Grounded-SAM-2~\cite{liu2024groundingdino,ravi2025sam2} for real-world data.
Frozen VGGT and Track4World targets are cached offline; RoboCasa motion
targets use the current frame and ten future frames uniformly sampled
over the 50-step chunk.

Training uses eight NVIDIA A100 80GB GPUs and AdamW
($\beta_1=0.9$, $\beta_2=0.95$, $\epsilon=10^{-8}$,
weight decay $10^{-10}$, gradient-norm clipping at 1.0).
Simulation training uses 30,000 updates with global batch sizes of 256
for LIBERO and 128 for RoboCasa; real-world training uses 16,000 updates
with a batch size of 256.  We linearly warm up to $5\times10^{-5}$ over
10,000 updates in simulation and 5,200 in real-world experiments.
The learning rate then remains constant, except on RoboCasa, where it
decays to $5\times10^{-6}$ with a cosine schedule.
For the full model, we use $\lambda_g=0.05$, $\lambda_s=0.01$, and
$\lambda_m=0.05$, with ablation-specific changes described below.
We use the official $\pi_{0.5}$ training and evaluation seeds.

\paragraph{Baselines}
We evaluate $\pi_{0.5}$ on RoboCasa using the same training and evaluation
settings as FOCAL-VLA. Results for the other simulation baselines are taken
from the cited sources.
GR00T N1.7 results follow the preliminary official LeRobot
evaluation~\cite{lerobot2026groot17}.  RoboCasa results for KYC, GeoVLA,
and PointVLA use Pointmap's $\pi_{0.5}$-based
reimplementations~\cite{lee2026pointmap}; Being-H0.5 uses its
RoboCasa-specific model.

\subsection{Evaluation Results}
\label{sec:exp_libero}

\paragraph{LIBERO}
Table~\ref{tab:libero} summarizes results on LIBERO.  FOCAL-VLA achieves an
average success rate of 97.9\%, improving over $\pi_{0.5}$ by 1.0 percentage
point.  It reaches 100.0\% on Spatial and obtains its largest gain on Long,
improving from 92.4\% to 95.0\%.  These results are consistent with our
joint modeling of current geometry and future interaction dynamics:
geometry distillation emphasizes the spatial
relationships needed for the current operation, while implicit world modeling
complements this information with anticipated changes in the interacting
entities.  As a task progresses, the supervision follows the entities involved
in each subtask rather than treating all image regions equally.

\paragraph{RoboCasa}
Table~\ref{tab:robocasa} reports results on RoboCasa.  FOCAL-VLA achieves
63.4\% average success, improving over $\pi_{0.5}$ by 8.2 percentage points
and outperforming the other listed methods.  Kitchen manipulation requires
selecting the objects and fixtures involved in the current operation from a
cluttered scene.  Subtask-guided geometry distillation targets the spatial
structure of these entities, while implicit world modeling captures their
future interaction dynamics using features from current and future
demonstration frames.  The two
branches therefore provide complementary information about the same
interaction: its current configuration and its future evolution.  This design
offers an explanation for the improvement over the base policy, with the
ablations below examining the roles of both distillation branches,
supervision scope, and action conditioning.

\begin{table}[t]
    \centering
    \caption{Comparison on RoboCasa.
    The best success rate is shown in bold.}
    \label{tab:robocasa}
    \small
    \setlength{\tabcolsep}{6pt}
    \begin{tabular*}{\columnwidth}{@{\extracolsep{\fill}}lc@{}}
        \toprule
        Method & Avg. SR (\%) \\
        \midrule
        \multicolumn{2}{@{}l}{\textbf{Base VLAs}} \\
        $\pi_0$~\cite{kim2026pri4r} & 38.8 \\
        $\pi_{0.5}$~\cite{black2025pi05} & 55.2 \\
        OpenVLA-OFT~\cite{kim2026pri4r} & 33.1 \\
        Being-H0.5~\cite{luo2026beingh05} & 53.9 \\
        \midrule
        \multicolumn{2}{@{}l}{\textbf{3D inputs required at inference}} \\
        KYC~\cite{jiang2026kyc,lee2026pointmap} & 59.1 \\
        GeoVLA~\cite{sun2025geovla,lee2026pointmap} & 57.1 \\
        PointVLA~\cite{li2025pointvla,lee2026pointmap} & 57.3 \\
        Robot-Centric Pointmap~\cite{lee2026pointmap} & 62.9 \\
        \midrule
        \multicolumn{2}{@{}l}{\textbf{3D inputs not required at inference}} \\
        GeoPredict~\cite{qian2026geopredict} & 52.4 \\
        GaussianDream~\cite{zhang2026gaussiandream} & 54.8 \\
        Pri4R~\cite{kim2026pri4r} & 57.0 \\
        \textbf{FOCAL-VLA} & \textbf{63.4} \\
        \bottomrule
    \end{tabular*}
\end{table}

\paragraph{Real-world experiments}
\label{sec:exp_real}

Table~\ref{tab:real_world} reports the real-world results on the four tasks
illustrated in Fig.~\ref{fig:real_world_tasks}.  FOCAL-VLA improves success
on all four tasks, raising the overall success rate from 47.5\% to 63.8\%.
Two-fruit rearrangement shows the largest gain and requires switching
between instructed objects and destinations.  Its improvement provides
further evidence of the method's effectiveness in sequential manipulation.
Drawer storage similarly requires the robot to place the banana inside the drawer
and then switch to closing it.
Completing this sequence involves a change in both the interacting entity
and the required motion.  The gain is consistent with our motivation to
associate geometric and future motion information with the current subtask:
the policy must use the banana--drawer spatial relationship during placement
and the drawer's subsequent motion during closing.
Gains on Drawer opening and Close the microwave
indicate that the benefits also extend to individual articulated-object
interactions, rather than being limited to tasks with multiple subtasks.
Some real-world grasping failures occur when the target object is occluded
in the single fixed camera view.

\begin{table}[t]
    \centering
    \caption{Real-world task success rates (\%).  Each method is evaluated
    in 20 trials per task.}
    \label{tab:real_world}
    \small
    \setlength{\tabcolsep}{6pt}
    \begin{tabular}{lcc}
        \toprule
        Task & $\pi_{0.5}$ & FOCAL-VLA \\
        \midrule
        Two-fruit rearrangement & 25.0 & \textbf{50.0} \\
        Drawer opening & 70.0 & \textbf{80.0} \\
        Drawer storage & 20.0 & \textbf{40.0} \\
        Close the microwave & 75.0 & \textbf{85.0} \\
        \midrule
        Overall & 47.5 & \textbf{63.8} \\
        \bottomrule
    \end{tabular}
\end{table}

\subsection{Ablation Studies}
\label{sec:exp_ablations}

We ablate the two distillation branches, supervision scope, and action
conditioning on RoboCasa under the same training and evaluation protocol
(Table~\ref{tab:ablations}).

\begin{table}[t]
    \centering
    \caption{Ablations on RoboCasa.}
    \label{tab:ablations}
    \small
    \setlength{\tabcolsep}{5pt}
    \begin{tabular}{@{}lc@{}}
        \toprule
        Variant & Success (\%) \\
        \midrule
        Full method & \textbf{63.4} \\
        \midrule
        Without geometry & 60.0 \\
        Without future motion & 59.6 \\
        Task-level entity supervision & 61.9 \\
        Full-image supervision & 58.7 \\
        Supervision-only & 58.5 \\
        \bottomrule
    \end{tabular}
\end{table}

\paragraph{Geometry and future motion}
To evaluate each teacher's contribution, we separately remove the geometry
or future motion queries together with the corresponding distillation loss.
Removing geometry reduces
success from 63.4\% to 60.0\%, while removing future motion reduces it to
59.6\%.  The drop without future motion supports learning anticipated
interaction changes in addition to current geometry.  Conversely, the drop
without geometry shows the benefit of retaining a dedicated representation of
the current spatial configuration.  Together, these results support combining
subtask-relevant geometry distillation with implicit world modeling, rather
than relying on either learning signal alone.

\paragraph{Supervision scope}
To evaluate the scope of supervision, we compare subtask-relevant supervision
with task-level entity supervision and full-image supervision.
Different subtasks within a task may involve different entities.
Task-level entity supervision uses the union of entities relevant to all
subtasks throughout the task, whereas subtask-level supervision selects only
those relevant to the current subtask.  In both cases, masks follow the
entities in each frame.  Both geometry and future motion distillation use
these masks, with all other settings unchanged.  Full-image supervision instead
pools VGGT and Track4World features over the entire
image~\cite{li2026spatialforcing,guo2025glad,wang2026track4action}.
Task-level entity supervision achieves 61.9\% success, exceeding full-image
supervision at 58.7\%, while the full model achieves 63.4\%.
These results suggest that selecting task-relevant entities improves
performance, with a further 1.5-percentage-point gain from adapting the
supervised entities to the current subtask.

\paragraph{Action conditioning}
To evaluate the benefit of direct action conditioning, we retain both
distillation branches but disconnect their output latents from the action
expert, yielding a supervision-only variant.  The supervision-only variant
achieves 58.5\% success, compared with
63.4\% for the full model.  This 4.9-percentage-point gain supports directly
conditioning the action expert on the learned latents, rather than using
them solely as auxiliary training signals.

\section{Limitations}
\label{sec:limitations}

FOCAL-VLA relies on offline subtask annotation and teacher-target construction,
which add data-preparation and computational costs.  Errors in subtask
boundaries or entity masks can affect the quality of the distillation targets.
Our evaluation covers two simulation benchmarks and a single real-world robot
platform, leaving generalization to other embodiments and task families
unexplored.  Future work will investigate reducing the annotation effort and
extending evaluation to more diverse robotic settings.

\section{Conclusion}
\label{sec:conclusion}

We presented FOCAL-VLA, a framework for enhancing VLA models through
subtask-guided geometry distillation and implicit world modeling.  Geometry distillation
aligns the policy's geometry latents with VGGT features from entities relevant
to each subtask.  Implicit world modeling is implemented through future-informed
motion distillation, using Track4World features from current and future
demonstration frames.  These features are pooled over subtask-relevant regions
to construct the distillation targets.
The two groups of latents jointly
guide action generation without requiring teacher models or subtask annotations
at inference time.  Experiments on LIBERO, RoboCasa, and real-world manipulation
tasks show higher average success rates than the $\pi_{0.5}$ baseline.
These results highlight the value of subtask-relevant geometric and future
motion knowledge for improving VLA manipulation performance.

\bibliographystyle{IEEEtran}
\bibliography{references}

\end{document}